# Research on Disease Prediction Model Construction Based on Computer AI deep Learning Technology

Yang Lin[1], Muqing Li[2], Ziyi Zhu[3], Yinqiu Feng[4], Lingxi Xiao[5], Zexi Chen[6]

[1]University of Pennsylvania,USA,yang.lin1345@gmail.com

[2]University of California San Diego,USA,MUL003@ucsd.edu

[3]New York University,USA,zz1831@nyu.edu

[4]Columbia University,USA,yf2579@columbia.edu

[5]Georgia Institute of Technology,USA,Lingxi.xiao@gatech.edu

[6]North Carolina State University, USA,chenzx.hit09@gmail.com

*Abstract*—The prediction of disease risk factors can screen vulnerable groups for effective prevention and treatment, so as to reduce their morbidity and mortality. Machine learning has a great demand for high-quality labeling information, and labeling noise in medical big data poses a great challenge to efficient disease risk warning methods. Therefore, this project intends to study the robust learning algorithm and apply it to the early warning of infectious disease risk. A dynamic truncated loss model is proposed, which combines the traditional mutual entropy implicit weight feature with the mean variation feature. It is robust to label noise. A lower bound on training loss is constructed, and a method based on sampling rate is proposed to reduce the gradient of suspected samples to reduce the influence of noise on training results. The effectiveness of this method under different types of noise was verified by using a stroke screening data set as an example. This method enables robust learning of data containing label noise.

## I. Introduction

Disease risk prediction is a prediction of the likelihood that people with certain characteristics will develop certain diseases in the future. The prediction of risk factors can help patients realize their own risk of disease and take early prevention and control actions, thus reducing their morbidity and mortality. At present, the risk prediction method of infectious diseases based on machine learning has been widely adopted and plays an important auxiliary role in medical services [1]. However, because of its precise dependence on the marking information, its performance is greatly restricted. However, medical big data is often manually completed by physicians, so it is inevitable that some false notes will appear in a large number of medical big data. When the accuracy of the label is not high enough, it will greatly reduce the efficiency of machine learning[2].

Because of the inaccuracy of the labeling information, there is the actual data, which brings noise to the training of the model [3]. This is mainly due to: (1) labeling bias, because each physician's diagnosis and treatment experience is different, the same medical data labeling will also have a large deviation. (2) The amount of test data in the TCM labeling (diagnosis and treatment) stage is insufficient, resulting in the characteristics of the sample not accurately characterizing the corresponding markers; (3) The recognition degree of labeled samples is not high; (4) Problems with the coding or communication of the data stored in medical data. The learning algorithm with label noise will degrade the performance of the algorithm, fail to achieve the expected effect, and require a large number of training samples, and the modeling complexity is high. At present, many researchers have conducted in-depth research on the problem of labeled noise, including sampling separation method, lost function-based method, sampling reweighting method, and so on. By filtering the sampling points containing noise, the algorithm avoids interference with the marked noise. Some studies have shown that in the training of neural networks, there is a problem of first having pure samples and then having noisy samples [4]. Some scholars have proposed a new method of constructing unlabeled noise based on low lost samples at training cost. Empirical evidence demonstrates that a multitude of networks exhibit distinct characteristics, enabling them to efficaciously discriminate and filter various categories of marker noise. This capability is vital to the objectives of this project, as it contributes significantly to the robustness and accuracy of the resultant analytical models[5]. They studied a collaborative learning algorithm based on CNN, that is, a small number of lost samples are selected as training objects in the same network, and their resistance to labeling noise is improved through mutual supervision. Some researchers have proposed a symmetric mutual information loss function, and applied it to the samples with noise marks, so as to effectively overcome the problems of poor learning ability and overfitting of cross-entropy loss function in the case of noise marks [6]. Some researchers have introduced the Taylor cross-entropy loss function and used the Taylor sequence to weigh the matching degree of the markers, so as to enhance the anti-noise ability of the markers. By assigning different weights to each sample, the algorithm adjusts its gradient and reduces the weights of potentially contaminated samples[7]. Previous studies have used Euclidean distance as the probability distribution of the initial sample, partitioned the sample, and designed the corresponding detection and filtering criteria to reduce the marking noise. Combined with the existing training sample information, the purpose of directional denoising is achieved through local design [8]. Previous studies have proposed using an unconstrained LLS importance method to evaluate the importance of labels, and integrating it with an autonomous learning method to complete the re-weighting of samples. Although a variety of algorithms have achieved good results in image recognition, there are few studies on disease risk prediction based on marked noise [9]. This project intends to study the robust learning algorithm based on labeled noise, and integrate the robust loss function with the sampling re-weighting technology to build an asymmetric bounded loss function to enhance the robustness against labeled noise [10]. By dynamically estimating the weights of each sampling point, the effect of label noise on disease risk prediction can be effectively reduced.

## II. ADAPTIVE MODULAR NEURAL NETWORK

### A. Structure principle of artificial neural network

This model is different from the common BP neural network in that the former preprocesses the data first, and then classifies and researches the problems in a modular way. AMNN is composed of multiple functions [11]. Its working principle is as follows: After cluster analysis of each training sample, it is divided into several categories, and according to the clustering results, the corresponding model in AMNN is selected for learning. The algorithm can realize the learning of the selected sample by clustering the sample set and constructing the selected sample according to the selected sample [12]. AMNN performs cluster analysis on the data, determines the cluster center, selects the subnetwork for learning according to the clustering results, and the subnetwork adaptively learns according to the obtained subnetwork. The network structure of AMNN is shown in Figure 1.

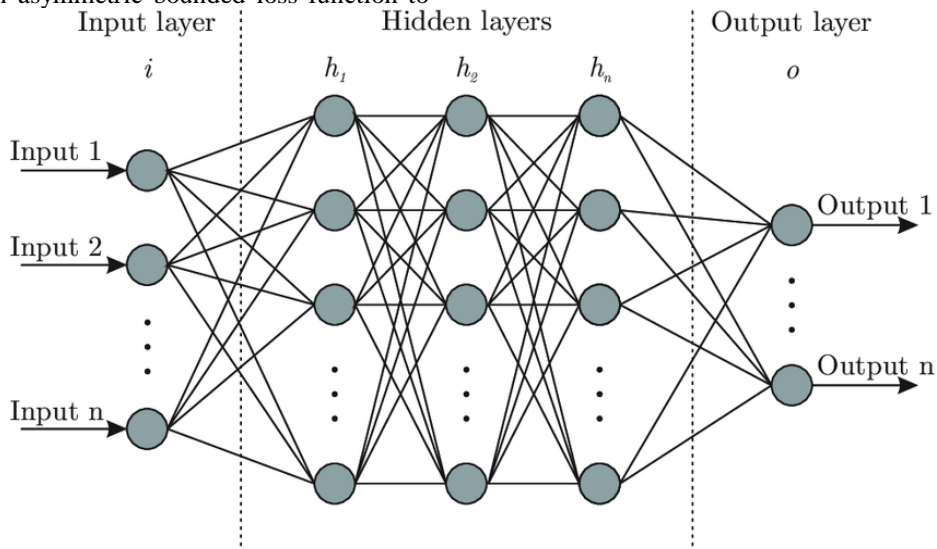

Fig. 1. Artificial neural network structure

### B. Task Decomposition

After completing the classification of the data set, it is classified, and it is classified. Input the newly generated sampling points into the corresponding subnet. The number of subsample Spaces in this algorithm is consistent with the number of subnets [13]. This method can judge the number of cluster centers of training sample clusters, and then determine the number of subnets. First, people want to ensure that the local density of each cluster's "neighbors" does not exceed the local density of the cluster. The core of the second cluster is well separated from other data with higher local intensity. Two parameters can be used to illustrate the principle of point density peak cluster algorithm. The local probability density of the data point i is labeled $\alpha_i$, and $\beta_i$ is distance from i to local density. Suppose A training sample set is $R = \{(u_t, v_t), t = 1, 2, \cdots, N\}$, then any data samples $\alpha_i$ and $\beta_i$ in $R$ are defined as follows:

$$\alpha_i = \sum_{j=1, j \neq i}^{N} \exp-\left(\frac{l_{ij}}{l_z}\right)^2 \quad (1)$$

$$\beta_i = \min_{j:\alpha_j > \alpha_i} l_{ij} \quad (2)$$

Where $l_{ij} = dist(u_i, u_j)$ represents the distance between the two sampling points $u_i$ and $u_j$, and this distance is expressed by the European distance, and $l_z$ is the truncated length, which has a value greater than 0. The choice of $l_z$ is directly related to the calculation effect of the method. If it is too large, the value of $\alpha_i$ may be too large and difficult to distinguish [14]. In the limit case, all the values are set together. If the selection is too small, it will cause a group of two groups or even multiple groups, so the selection of group $l_z$ is very critical. In this paper, the spacing $l_{ij}$ between the two groups is arranged in ascending order, the largest 2% is selected and rounded to get $l_z$, thus reducing the dependence on specific problems to a certain extent.

The center of the data aggregation category is $\{z_1, z_2, \cdots, z_G\}$. Finally, the aggregation center of cluster $G$ is formed [15]. The time complexity is $P(n^2)$. For the center of each cluster, a fuzzy set is generated for all sampling points in a set, resulting in a fuzzy set correlation (3). The sum of $G$ fuzzy sets is expressed as follows:

$$g_{ik} = \exp\left(\frac{-\|u_t - z_i\|^2}{0.02}\right) \quad (3)$$

Where, $g_{ik}$ is a fuzzy membership for subordinating training sample $u_t$ to the $i$ fuzzy group. The index partial factor represents $u_t$ and $z_i$, and the denominator is chosen to increase its value to improve its recognition rate [16]. If it is too large, it will lead to its ownership distinction is not uniform, and if it is too small, it will lead to its "sharp" and "steep", resulting in it approaching 0, and then it will be difficult to distinguish its ownership. 02 is the best choice [17]. For each sampling point, if $u_t$ is close to the cluster center $z_i$, $u_t$ has a high degree of membership to $z_i$, and $u_t$ is divided into the corresponding subsample space as the input subnetwork for training.

### C. Determine the learning parameters of the neural network adaptively

Reverse transmission of the reverse transmission error and the reverse transmission of forward transmission is carried out, and then layers of regression, and the connection weight of each level are corrected according to the size of the error[18]. The above steps are described as follows. When the actual output $\overline{v}_t$ of the output layer does not match the expected output $v_t$, the error function is:

$$W = \frac{1}{2}\sum_K (v_t - \overline{v}_t)^2 \quad (4)$$

The modification of the join weight is calculated according to formula (4).

$$\lambda_{jk}(k+1) = \lambda_{jk}(k) + \Delta\lambda_{jk} \quad (5)$$

BP algorithm uses the gradient descent direction to modify the connection weight, and the weight change is

$$\Delta\lambda_{jk} = -\zeta\frac{\partial W}{\partial \lambda_{jk}} = -\zeta\varepsilon_t P_j \quad (6)$$

Where the learning rate $\zeta$ is $0 < \zeta < 1, \varepsilon_t = \frac{\partial W}{\partial \overline{v}_t}\frac{\partial \overline{v}_t}{\partial \sigma_t}, \sigma_t$, representing the input of the $t$ neuron in the output layer, and $P_j$, representing the output of the $j$ neuron in the hidden layer

$$\Delta\lambda_{jk} = \zeta\overline{v}_t(1-\overline{v}_t)(v_t - \overline{v}_t)P_j \quad (7)$$

For the hidden layer there is

$$\Delta v_{ij} = \zeta P_j(1-P_j)\sum_t \varepsilon_t \lambda_{jk} X_i \quad (8)$$

The learning and training of BP neural network is a repetitive cycle, that is, the weights at each level are corrected through positive and negative feedback until the final result meets the required accuracy or the maximum number of trainings [19].

### III. EXPERIMENTAL RESULTS AND ANALYSIS

The various types of marked noise conditions is proved by an example analysis of stroke screening experiments [20]. In this study, the risk of stroke was divided into three levels: Level 1 represented low-risk individuals, who had a relatively high risk of developing the disease, while low-risk individuals required routine physical examination [21]. The second level is medium risk, belonging to the high-risk group, which requires routine health checks and corresponding management of high-risk factors. Level 3 indicates a high risk and requires a detailed diagnosis in a regular hospital, followed by treatment under the guidance of a specialist [22]. The proportions of these three stroke risk levels in the data set are shown in Table 1. The reference markers were rearranged in the experiment, and the specified markers were replaced according to a specific ratio, in the range of 5% to 40%.

TABLE I. LABEL DISTRIBUTION OF THE DATASET

| Stroke risk level | quantity | Proportion (%) |
|---|---|---|
| low-risk | 638353 | 74.03 |
| Moderately dangerous | 129274 | 14.99 |
| High risk | 130544 | 15.15 |

This article will use the Python languages PyTorch and Scikit-learn to complete DNN and dynamic truncation methods [23]. The hyperparameter Settings that are relevant to the test are shown in Table 2. All other parameters not mentioned are default.

TABLE II. MODEL HYPERPARAMETERS

| Module | Argument | Settings |
|---|---|---|

| Model | Layer size | 20 |
|---|---|---|
| | Activation function | ELU |
| | q | 0.7 |
| | k | 0.5 |
| Optimizer | Test set ratio | 0.2 |
| | Optimization algorithm | Adam |
| | Batch size | 128 |
| | Learning rate | 1 x 10-4 |
| | Learning cycle number | 20 |

This paper will compare stroke risk with marked noise. Aiming at stroke risk factors that are not marked with noise, this project intends to use deep neural network, decision tree, random forests, LightGBM, XGBoost, CatBoost, and other methods to assess stroke risk [24]. Decision tree method has been applied more and more in medical diagnosis and prediction [25]. Random forests are an efficient model for predicting disease risk. LightGBM has been used more and more in medicine. XGBoost technology has been greatly developed in clinical practice, and has been well applied in the fields of early warning and risk prediction of diseases [26]. This algorithm can achieve good performance prediction without any optimization, and has a good application prospect in medical diagnosis [27]. This paper will use deep neural network, LightGBM, XGBoost, CatBoost, DNN tracking, DNN mixing, and other methods to study. Mixup is a class of data enhancement methods that reduce the gradient of contaminated samples by smoothing features and noisy labels to suppress.

In the absence of marked noise, this paper will use decision trees, random forests, LightGBM, XGBoost, CatBoost, and deep neural networks to achieve the comparison of these methods [28]. It can be seen from data in Table 3 that the accuracy rate of the above methods is above 94%, which meets the practical requirements, while the performance of decision tree and random forest methods is poor, while the accuracy rate, weighted accuracy rate and weighted recall rate of LightGBM, XGBoost, CatBoost, and DNN methods are above 98.70%, respectively. The weighted F1 score was greater than 0.9870, and the Kappa coefficient was greater than 0.9710, which could better identify various risk levels and achieve better classification effects (Table 3). This is probably because the single tree structure has poor predictive performance and is easily disturbed by outliers without adopting an integrated learning method[29]. The comprehensive learning method is based on the conclusion of multiple trees to reduce the influence of outliers on system performance. Random forest adopts the method of random sampling and random selection to construct multiple trees, LightGBM, XGBoost, CatBoost, etc., all adopt the method of gradual increase to train, so that it has higher accuracy. According to the above tests, LightGBM, XGBoost, CatBoost, DNN and other methods can be used for early diagnosis of stroke (Table 4).

TABLE III. COMPARISON OF TEST SET ACCURACY WITHOUT LABEL NOISE

| Algorithm | Accuracy rate (%) | Weighted accuracy (%) | Weighted recall rate (%) | Weighted F1 score | Kappa coefficient |
|---|---|---|---|---|---|
| Decision tree | 95.364 | 95.212 | 95.364 | 0.949 | 0.885 |
| Random forest | 97.162 | 97.091 | 97.162 | 0.969 | 0.925 |
| LightGBM | 99.747 | 99.889 | 99.747 | 0.998 | 0.983 |
| XGBoost | 99.727 | 99.869 | 99.727 | 0.998 | 0.983 |
| CatBoost | 99.727 | 99.869 | 99.727 | 0.997 | 0.983 |
| DNN | 99.697 | 99.808 | 99.697 | 0.997 | 0.981 |

TABLE IV. COMPARISON OF TEST SET ACCURACY UNDER DIFFERENT NOISE RATIOS (%)

| Algorithm | 5 | 10 | 15 | 20 | 25 | 30 | 35 | 40 |
|---|---|---|---|---|---|---|---|---|
| DNN | 97.41 | 95.00 | 92.56 | 90.22 | 87.88 | 85.43 | 83.03 | 80.90 |
| LightGBM | 97.12 | 94.56 | 91.90 | 89.16 | 86.61 | 84.14 | 81.41 | 78.82 |
| XGBoost | 97.16 | 94.45 | 91.93 | 89.22 | 86.52 | 84.17 | 81.52 | 78.69 |
| CatBoost | 97.10 | 94.53 | 91.99 | 89.31 | 86.51 | 84.07 | 81.37 | 78.80 |
| DNN-mixup | 78.02 | 78.33 | 77.76 | 77.69 | 77.77 | 77.01 | 76.89 | 75.81 |
| Textual algorithm | 97 | 97 | 97 | 97 | 97 | 97 | 97 | 97 |

## IV. CONCLUSION

The study presents a robust framework for disease risk prediction leveraging artificial intelligence and machine learning technologies. It introduces a dynamic truncated loss model that effectively handles label noise, enhancing the accuracy of disease prediction models. The research showcases how the integration of machine learning methods like deep neural networks, decision trees, and ensemble methods such as LightGBM. XGBeost, and CatBoost can significantly improve the precision of identifying risk levels in disease prediction. Experimental results, particularly using stroke risk data, illustrate the practicality and efficiency of the proposed methods under various conditions of label noise, confirming the robustness of the approach. Overall, this research advances the field of medical diagnostics and offers a promising direction for future exploration to further refine and validate the proposed methods on a broader scale.

REFERENCES

[1] Praskatama, V., Sari, C. A., Rachmawanto, E. H., & Yaacob, N. M. PNEUMONIA PREDICTION USING CONVOLUTIONAL NEURAL NETWORK. Jurnal Teknik Informatika (Jutif), vol. 4, pp.1217-1226, May 2023.

[2] Dai, W., Tao, J., Yan, X., Feng, Z., & Chen, J. (2023, November). Addressing Unintended Bias in Toxicity Detection: An LSTM and Attention-Based Approach. In 2023 5th International Conference on

Artificial Intelligence and Computer Applications (ICAICA) (pp. 375-379). IEEE.

[3] Hu, Z., Li, J., Pan, Z., Zhou, S., Yang, L., Ding, C., ... & Jiang, W. (2022, October). On the design of quantum graph convolutional neural network in the nisq-era and beyond. In 2022 IEEE 40th International Conference on Computer Design (ICCD) (pp. 290-297). IEEE.

[4] Mann, P. S., Panchal, S. D., Singh, S., Saggu, G. S., & Gupta, K. A hybrid deep convolutional neural network model for improved diagnosis of pneumonia. Neural Computing and Applications, vol. 36, pp.1791-1804, April 2024.

[5] Yan, X., Wang, W., Xiao, M., Li, Y., & Gao, M. (2024). Survival Prediction Across Diverse Cancer Types Using Neural Networks. arXiv preprint arXiv:2404.08713.

[6] Liu, Z., Yang, Y., Pan, Z., Sharma, A., Hasan, A., Ding, C., ... & Geng, T. (2023, July). Ising-cf: A pathbreaking collaborative filtering method through efficient ising machine learning. In 2023 60th ACM/IEEE Design Automation Conference (DAC) (pp. 1-6). IEEE.

[7] Wang, X. S., Turner, J. D., & Mann, B. P. (2021). Constrained attractor selection using deep reinforcement learning. Journal of Vibration and Control, 27(5-6), 502-514.

[8] Yao, J., Li, C., Sun, K., Cai, Y., Li, H., Ouyang, W., & Li, H. (2023, October). Ndc-scene: Boost monocular 3d semantic scene completion in normalized device coordinates space. In 2023 IEEE/CVF International Conference on Computer Vision (ICCV) (pp. 9421-9431). IEEE Computer Society.

[9] Xin Chen , Yuxiang Hu, Ting Xu, Haowei Yang, Tong Wu. (2024). Advancements in AI for Oncology: Developing an Enhanced YOLOv5-based Cancer Cell Detection System. International Journal of Innovative Research in Computer Science and Technology (IJIRCST), 12(2),75-80, doi:10.55524/ijircst.2024.12.2.13.

[10] Atulya Shree, Kai Jia, Zhiyao Xiong, Siu Fai Chow, Raymond Phan, Panfeng Li, & Domenico Curro. (2022). Image analysis.

[11] Guo, A., Hao, Y., Wu, C., Haghi, P., Pan, Z., Si, M., ... & Geng, T. (2023, June). Software-hardware co-design of heterogeneous SmartNIC system for recommendation models inference and training. In Proceedings of the 37th International Conference on Supercomputing (pp. 336-347).

[12] Wang, X. S., & Mann, B. P. (2020). Attractor Selection in Nonlinear Energy Harvesting Using Deep Reinforcement Learning. arXiv preprint arXiv:2010.01255.

[13] Kijowski, R., Liu, F., Caliva, F., & Pedoia, V. Deep learning for lesion detection, progression, and prediction of musculoskeletal disease. Journal of magnetic resonance imaging, vol. 52, pp.1607-1619, June 2020.

[14] Babukarthik, R. G., Adiga, V. A. K., Sambasivam, G., Chandramohan, D., & Amudhavel, J. Prediction of COVID-19 using genetic deep learning convolutional neural network (GDCNN). Ieee Access, vol. 8, pp.177647-177666, May 2020.

[15] Katarya, R., & Meena, S. K. Machine learning techniques for heart disease prediction: a comparative study and analysis. Health and Technology, vol. 11, pp.87-97, January 2021.

[16] Rasmy, L., Xiang, Y., Xie, Z., Tao, C., & Zhi, D. Med-BERT: pretrained contextualized embeddings on large-scale structured electronic health records for disease prediction. NPJ digital medicine, vol. 4, pp.86-89, January 2021.

[17] Khan, P., Kader, M. F., Islam, S. R., Rahman, A. B., Kamal, M. S., Toha, M. U., & Kwak, K. S. Machine learning and deep learning approaches for brain disease diagnosis: principles and recent advances. Ieee Access, vol. 9, pp. 37622-37655, August 2021.

[18] Zhang, K., Liu, X., Xu, J., Yuan, J., Cai, W., Chen, T., ... & Wang, G. Deep-learning models for the detection and incidence prediction of chronic kidney disease and type 2 diabetes from retinal fundus images. Nature biomedical engineering, vol. 5, pp. 533-545, June 2021.

[19] Rani, P., Kumar, R., Ahmed, N. M. S., & Jain, A. A decision support system for heart disease prediction based upon machine learning. Journal of Reliable Intelligent Environments, vol. 7, pp. 263-275, March 2021.

[20] Swathy, M., & Saruladha, K. A comparative study of classification and prediction of Cardio-Vascular Diseases (CVD) using Machine Learning and Deep Learning techniques. ICT Express, vol. 8, pp. 109-116, January 2022.

[21] Zeng, M., Lu, C., Zhang, F., Li, Y., Wu, F. X., Li, Y., & Li, M. SDLDA: lncRNA-disease association prediction based on singular value decomposition and deep learning. Methods, vol. 179, pp. 73-80, November 2020.

[22] Yao, J., Wu, T., & Zhang, X. (2023). Improving depth gradient continuity in transformers: A comparative study on monocular depth estimation with cnn. arXiv preprint arXiv:2308.08333.

[23] Yufeng Li, Weimin Wang, Xu Yan, Min Gao, & MingXuan Xiao. (2024). Research on the Application of Semantic Network in Disease Diagnosis Prompts Based on Medical Corpus. International Journal of Innovative Research in Computer Science & Technology, 12(2), 1–9. Retrieved from https://ijircst.irpublications.org/index.php/ijircst/article/view/29

[24] Yulu Gong , Haoxin Zhang, Ruilin Xu, Zhou Yu, Jingbo Zhang. (2024). Innovative Deep Learning Methods for Precancerous Lesion Detection. International Journal of Innovative Research in Computer Science and Technology (IJIRCST), 12(2),81-86, doi:10.55524/ijircst.2024.12.2.14.

[25] Zi, Y., Wang, Q., Gao, Z., Cheng, X., & Mei, T. (2024). Research on the Application of Deep Learning in Medical Image Segmentation and 3D Reconstruction. Academic Journal of Science and Technology, 10(2), 8-12.

[26] Weimin WANG, Yufeng LI, Xu YAN, Mingxuan XIAO, & Min GAO. (2024). Enhancing Liver Segmentation: A Deep Learning Approach with EAS Feature Extraction and Multi-Scale Fusion. International Journal of Innovative Research in Computer Science & Technology, 12(1), 26–34. Retrieved from https://ijircst.irpublications.org/index.php/ijircst/article/view/21

[27] Dubey, A. K. Optimized hybrid learning for multi disease prediction enabled by lion with butterfly optimization algorithm. Sādhanā, vol. 46, pp.63-69, February 2021.

[28] Li, P., Abouelenien, M., & Mihalcea, R. (2023). Deception Detection from Linguistic and Physiological Data Streams Using Bimodal Convolutional Neural Networks. arXiv preprint arXiv:2311.10944.

[29] Zhicheng Ding, Panfeng Li, Qikai Yang, Xinyu Shen, Siyang Li, & Qingtian Gong (2024). Regional Style and Color Transfer. arXiv preprint arXiv:2404.13880.